\documentclass[lettersize,journal]{IEEEtran}
\usepackage{amsmath,amsfonts}
\usepackage{algorithmic}
\usepackage{algorithm}
\usepackage{array}
\usepackage[caption=false,font=normalsize,labelfont=sf,textfont=sf]{subfig}
\usepackage{textcomp}
\usepackage{stfloats}
\usepackage{url}
\usepackage{verbatim}
\usepackage{graphicx}
\usepackage{cite}
\usepackage{multirow}
\usepackage{booktabs}
\begin{document}

\title{Deep Convolutional Pooling Transformer for Deepfake Detection}

\author{Tianyi Wang, Harry Cheng, Kam Pui Chow, Liqiang Nie

\thanks{Tianyi Wang and Kam Pui Chow are with the Department of Computer Science, The University of Hong Kong, Hong Kong (e-mail: tywang@cs.hku.hk, chow@cs.hku.hk). }
\thanks{Harry Cheng is with the School of Computer Science and Technology, Shandong University, Qingdao, China (email: xaCheng1996@gmail.com).}
\thanks{Liqiang Nie is with the Department of Computer Science and Technology, Harbin Institute of Technology (Shenzhen), Shenzhen, China (email: nieliqiang@gmail.com).}
\thanks{Corresponding author: Kam Pui Chow.}
}



\maketitle

\begin{abstract}
Recently, Deepfake has drawn considerable public attention due to security and privacy concerns in social media digital forensics. As the wildly spreading Deepfake videos on the Internet become more realistic, traditional detection techniques have failed in distinguishing between real and fake. Most existing deep learning methods mainly focus on local features and relations within the face image using convolutional neural networks as a backbone. However, local features and relations are insufficient for model training to learn enough general information for Deepfake detection. Therefore, the existing Deepfake detection methods have reached a bottleneck to further improve the detection performance. To address this issue, we propose a deep convolutional Transformer to incorporate the decisive image features both locally and globally. Specifically, we apply convolutional pooling and re-attention to enrich the extracted features and enhance efficacy. Moreover, we employ the barely discussed image keyframes in model training for performance improvement and visualize the feature quantity gap between the key and normal image frames caused by video compression. We finally illustrate the transferability with extensive experiments on several Deepfake benchmark datasets. The proposed solution consistently outperforms several state-of-the-art baselines on both within- and cross-dataset experiments.
\end{abstract}


\section{Introduction}
\label{sec:intro}

\begin{figure}[t]
\centering
\includegraphics[width=0.9\columnwidth]{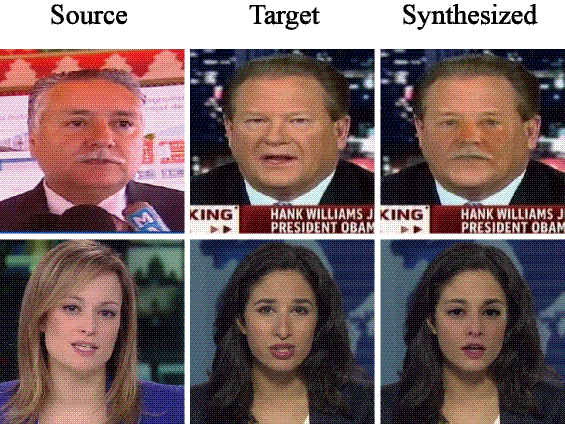} 
\caption{Examples of Deepfake. The source identity is swapped onto the target, maintaining the facial expression and action of the target person.}
\label{fig1}
\end{figure}

The fast-developing facial manipulation technique, Deepfake (examples are shown in Fig.~\ref{fig1}), has achieved remarkable success with various face synthesis algorithms in recent years~\cite{Kemelmacher2016Transfiguring, faceswap, Li2019FaceShifter, Natsume2018RSGAN}. While the high-quality Deepfake videos can facilitate human lives in multiple domains such as educational media and digital communications~\cite{Mika2019The}, they can be easily misused by criminals for malicious purposes and cause potential threats in societal, political, and business realms~\cite{Kirsti2019fake, Jan2020Deepfakes, Ruben2020Deepfakes}. Recognized as the most severe artificial intelligence threat in 2020~\cite{University2021Deepfakes}, Deepfake has been vastly adopted to generate fake celebrity porn videos and fake politician speeches~\cite{Hasan2019Combating}. Famous actress Emma Watson and the well-known singer Ariana Grande are representative victims of Deepfake porn videos~\cite{Leo2018Deepfake}. Moreover, the wildly spreading fake Barack Obama insulting Donald Trump video has reflected its severe social impact~\cite{obama2018video}. The number of Deepfake videos is growing at an unexpected rate such that it has almost doubled within only nine months in 2019~\cite{Rob2020Deepfakes}. Considering the free access to the Deepfake technique with open-source implementations and packaged Deepfake applications circulating on the Internet, anyone can become a victim of the Deepfake technique. Consequently, it is unrealistic to distinguish real and fake videos manually due to the vast amount of existing Deepfake videos and their increasing authenticity. Therefore, protecting human lives against Deepfake attacks with the help of automatic Deepfake detection solutions is highly desired to preserve the social order.

Deepfake detection is normally framed as a binary classification task by the existing approaches. While the classic forgery detection methods~\cite{Forensic2016Ravi, contrast2018wen} are insufficient to identify hyper-realistic Deepfake materials, deep learning models are frequently utilized. Specifically, the image frames are extracted from the candidate video and fed into the designed deep neural networks to determine the authenticity. Early approaches mainly focus on the local features in the input image to look for abnormal feature patches and resort to the pre-trained convolutional neural network (CNN) backbones followed by a fully connected layer to conduct the classification. In particular, the convolutional kernel studies information within each receptive field while ignoring the correlations between positions that are far away. Meanwhile, the goal of Deepfake detection has shifted from within-dataset to cross-dataset test on unseen Deepfake videos for generalization purposes. As the CNNs have shown poor generalization ability in feature learning, recent algorithms have gradually introduced the idea of the attention mechanism using CNNs to enlarge the local image feature areas for both within- and cross-dataset performance enhancement. However, recent synthetic work has gradually fulfilled the research gap and local artifacts are now difficult to be discovered. As finding consistency and relation between any two positions within an image can be meaningful such that the synthetic eyes and mouth may not look consistent with respect to the facial expressions, those current solutions are still limited in learning global features for further improving the Deepfake detection ability. On the other hand, the low resolution of the randomly extracted image frames has limited the potential ability of the detection models as they naturally contain restricted image features rather than high-resolution ones. 

To handle the aforementioned issues, we present a keyframe-based deep convolutional Transformer model, which jointly leverages convolutional pooling and re-attention approaches to study the decisive image features and relations both locally and globally. In detail, a stack of CNNs is adopted to extract local features from the input face images, and an enhanced deep Transformer with convolutional pooling and re-attention follows to enrich the global feature learning and to analyze the corresponding relations between image feature patches. Specifically, the convolutional pooling refines dominant features with dimension reduction, and the re-attention technique maintains the diversity within attention maps in a deep model. Moreover, since videos are commonly compressed for storage saving, unavoidable information loss occurs when the normal image frames are reconstructed and randomly extracted from the videos. On the contrary, the keyframes within a video are the only image frames that carry complete frame information with high resolution and do not suffer from information loss during the image frame reconstruction process~\cite{AnEfficient2017Dutta}. To the best of our knowledge, the importance of the keyframes has been untapped in the existing Deepfake detection approaches. In order to bridge this gap, in this study, we extract all keyframes from the videos for both training and testing processes, and thus emphasize the power of keyframes in boosting Deepfake detection performance. We conduct extensive experiments on several commonly-used Deepfake datasets. The experimental results prove that our approach outperforms several baselines in both within- and cross-dataset evaluations. We further visualize the gap in the extracted features between normal and key image frames to verify the robustness of the latter. 

The contributions of this work are threefold:
\begin{itemize}
\item We introduce the scheme of learning image features both locally and globally using the proposed deep convolutional Transformer model via jointly integrating convolutional pooling and re-attention strategies, which well explores the global features within a face image yet ignored by the existing methods.  
\item Significantly distinguished from the existing Deepfake detection solutions that seldomly consider the information loss in image extraction from the compressed videos, we are the first on extracting the keyframes and exhibiting their importance both statistically and visually in experiments, to the best of our knowledge.
\item Besides the considerable achievement of within-dataset performance, we verify the robustness of the proposed model in extensive cross-dataset experiments, which outperforms the existing state-of-the-art baselines on several popular Deepfake datasets. 
\end{itemize}

\section{Related Work}
\label{sec:related_work}

\subsection{Deepfake Generation}
\label{sec:rw_df_generation}

The term Deepfake is firstly introduced by the Reddit user `deepfakes' when the open-source implementation was released simultaneously in 2017. The leading architecture in the existing Deepfake generation work~\cite{faceswap, perov2021deepfacelab, Li2019FaceShifter} is an autoencoder~\cite{Kingma2014Autoencoder} for facial identity swap. Specifically, a shared encoder learns the identity-independent features from the input faces of different identities, and two individual decoders each generates synthesized faces with the desired identity. When performing face-swapping, a target face is passed into the autoencoder. The shared encoder performs facial feature extraction and converts the input target face to a context vector. Thenceforth, the corresponding decoder analyzes the context vector and generates a look maintaining the facial expression of the target face while having the identity of the desired source face. 

In light of the promising synthetic performance of Generative Adversarial Networks (GAN)~\cite{Goodfellow2014Generative} in domains such as face recognition~\cite{Zhao20183D, wang2021face, zhao2018towards}, face synthesis~\cite{Zhao2017Dual}, and human parsing~\cite{Zhao2018Understanding}, recent Deepfake research has exploited GAN for better synthesis authenticity. For example, FaceShifter~\cite{Li2019FaceShifter} and SimSwap~\cite{Chen2020SimSwap} both propose specific modules to preserve facial attributes that are hard to handle and maintain the fidelity for arbitrary identities. HifiFace~\cite{Wang2021HifiFace} and MegaFS~\cite{zhu2021megafs} achieve high-resolution identity swap on 512 and 1,024 resolutions for arbitrary identities, respectively.

\subsection{Deepfake Detection}
\label{sec:rw_df_detection}

Since the threat of Deepfake firstly attracted public attention, most existing Deepfake detection methods have utilized CNN-based architectures as the main backbones along with other fine-tuning steps. An early approach~\cite{Amerini2019Deepfake} proposes optical flow analysis with pre-trained VGG16~\cite{Karen2015Very} and ResNet50~\cite{He2016Deep} CNN backbones and achieves preliminary within-dataset test performance on the FaceForensics++ (FF++) dataset~\cite{Rossler2019FaceForensics}. Bonettini et al.~\cite{Bonettini2021Video} studied the combination of different pre-trained CNN models starting from EfficientNetB4~\cite{Tan2019EfficientNet} as the backbone for Deepfake detection. The DFT-MF~\cite{Jafar2020Forensics} approach focuses on open mouths with teeth and uses the standard CNN model to detect Deepfake by isolating, analyzing, and verifying lip and mouth movements. Later, Rossler et al.~\cite{Rossler2019FaceForensics} employed the pre-trained well-designed Xception~\cite{Chollet2017Xception} network and achieved the state-of-the-art detection performance over all other existing approaches at the time on the FF++ dataset. However, since the CNN architecture lacks generalization ability and mainly focuses on local features, even Xception is restricted in learning the global features for further performance improvements. Therefore, the most recent solutions have introduced the attention mechanism using a convolutional layer with the kernel size of one and claim to enlarge local feature areas and the corresponding relations. The SRM~\cite{Luo2021Generalizing} network constructs two streams of Xception backbones that focus on RGB frames and high-frequency frames, respectively, and studies the cross-modal relations between the two streams. The MAT model~\cite{Zhao2021Multi-Attentional} adopts EfficientNetB4 as the backbone to detect Deepfake and introduces the convolutional attention layer to study features within different local parts. Recently, unlike previous work that directly utilizes cross-entropy loss for real and fake classification, Sun et al.~\cite{sun2021dual} proposed Dual Contrastive Learning (DCL) to study positive and negative paired data for Deepfake detection with better transferability. It is also worth noting that a Transformer based approach, CViT~\cite{Deressa2021Deepfake}, has been proposed with source code published and has shown promising results, and several latest models~\cite{Khan2022Hybrid, Coccomini2022Combining, Coccomini2022MINTIME} are further designed following similar ideas but devoting better detection performance. 

\begin{figure*}[t]
\centering
\includegraphics[width=\textwidth]{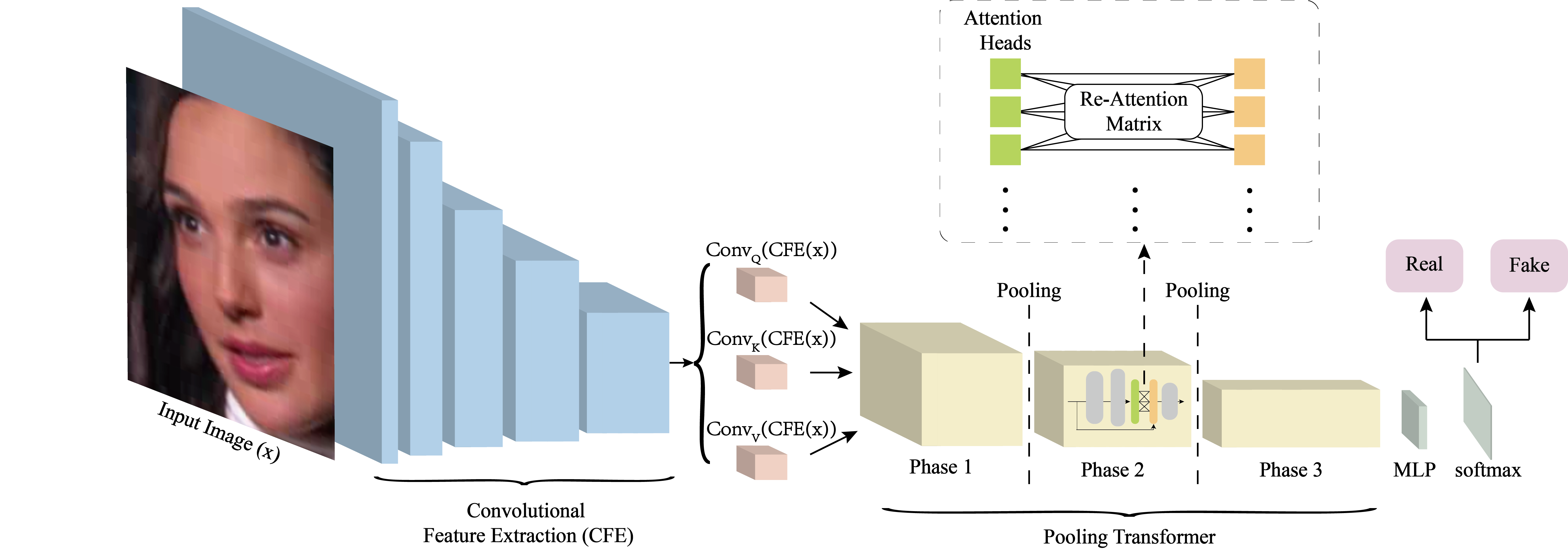}
\caption{Schematic illustration of the proposed model. Keyframe face images are passed through a stack of CNNs for feature extraction. The learned features are then projected as inputs by depth-wise separable convolutions and fed to the deep Transformer with convolutional pooling and re-attention. After that, the multilayer perceptron and softmax generate the final outputs.}
\label{fig2}
\end{figure*}

\subsection{Vision Transformers}
\label{sec:rw_vits}

The success of Transformer~\cite{Vaswani2017Attention} in Natural Language Processing (NLP) in 2017 has attracted researchers to explore its potential ability in the computer vision domain. Since the first vision Transformer (ViT)~\cite{Alexey2021An} achieves relatively reasonable performance on downstream vision tasks, various innovative vision Transformer architectures have been attempted. An early representative novel Swin Transformer~\cite{liu2021Swin} devises hierarchical architecture with shifted windows to make it compatible with a broad range of vision tasks. Later, a Contextual Transformer~\cite{Li2021Contextual} is proposed to learn contextual information among input keys that have been ignored by the classic self-attention. Yao et al.~\cite{Yao2022wavevit} conducted Wavelet Vision Transformer (Wave-ViT) following the wavelet theory, which specifically solves the inevitable information dropping caused by over-aggressive down-sampling when dealing with high computational costs. Recently, a Dual Vision Transformer (Dual-ViT)~\cite{Yao2022dualvit} further enhances the computational efficiency of self-attention with the state-of-the-art vision task performance by incorporating a critical semantic pathway that compresses token vectors into global semantics. Downstream tasks that satisfactory performance has been achieved by Transformer architectures include image recognition, object detection, and semantic segmentation. However, the promising idea of Transformer and self-attention has been barely discussed and attempted in the Deepfake detection domain. Therefore, in this study, we fulfill the research gap by presenting a deep convolutional Transformer model.

\section{Methodology}
\label{sec:methodology}

\subsection{Framework}
\label{sec:met_framework}

In this section, we briefly introduce the framework and workflow of the proposed approach. As shown in Fig.~\ref{fig2}, the model mainly consists of the following parts: convolutional feature extraction, depth-wise separable convolutional projection, and pooling Transformer. To be more specific, the face images are fed to the convolutional feature extraction module after being extracted and cropped from the candidate Deepfake image frames. The convolutional feature extraction module studies the local features within each face image and acquires the representative information. We then perform the depth-wise separable convolution upon the learned local feature information and transmit the results to the pooling Transformer. The pooling Transformer analyzes the global image features, and the attention map diversity is preserved by the re-attention mechanism. Particularly, the re-attention mechanism is applied to each attention map within the pooling Transformer module by unique learnable transformation matrices for each layer. Thereafter, the ultimately learned feature information is reshaped with multilayer perceptron and performed Deepfake detection.

In the following parts, we first introduce the idea of utilizing image keyframes within a commonly seen compressed video. We then illustrate the convolutional local feature extraction module upon the input images. After that, we demonstrate the Transformer part for global features and relation analysis. In particular, we present the idea of convolutional pooling and re-attention within an enhanced deep Transformer.

\subsection{Keyframe Extraction}
\label{sec:met_keyframe}

\begin{figure}
\centering
\includegraphics[width=\columnwidth]{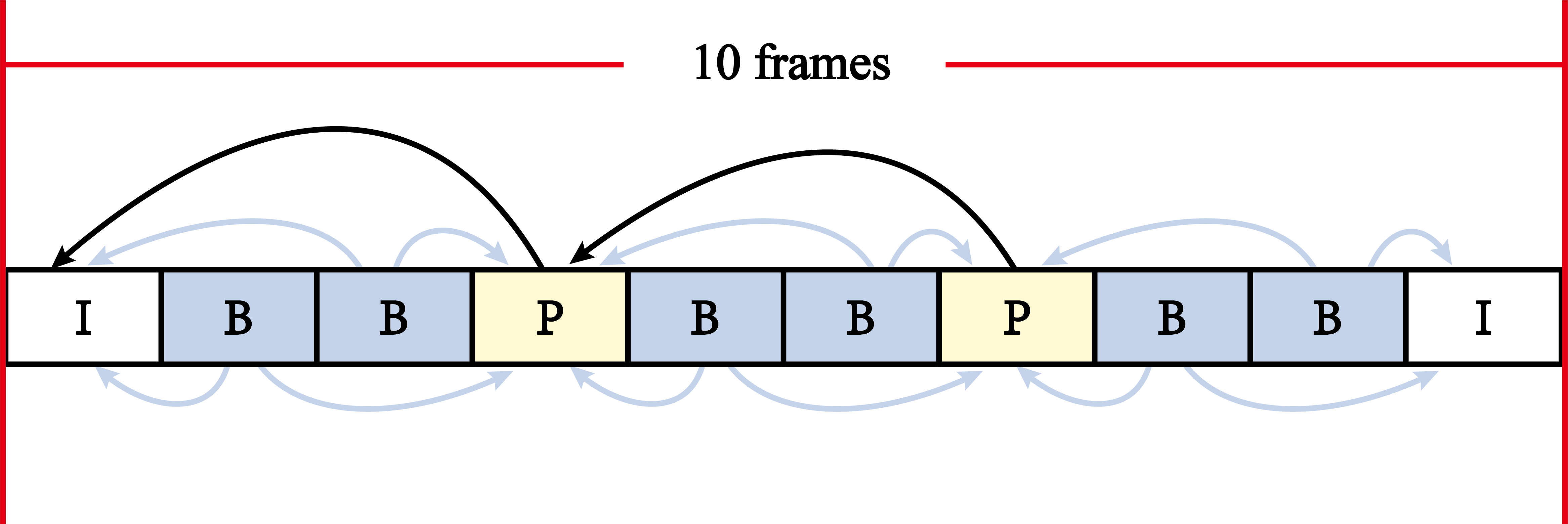}
\caption{A 10-frame H.264 compressed video clip demonstration showing frame dependencies. A B-frame relies on the closest succeeding and preceding I-frame or P-frame, and a P-frame is encoded depending on the closest preceding I-frame or P-frame, while an I-frame contains complete information of a particular image frame.}
\label{fig3}
\end{figure}

In real-life scenarios, video compression is usually operated to maintain as few complete image frames as possible for storage saving. The commonly used compression formats include H.264, MPEG-4, and JPEG~\cite{NORMAN2014203}. In general, three types of image frames, I-frame, P-frame, and B-frame, are derived within a video under H.264 compression. A 10-frame H.264 compressed video clip example showing the dependencies among the frames is exhibited in Fig.~\ref{fig3}. I-frame, also known as keyframe or intra-frame, is the only type of image frame that carries complete image information with the largest sizes and resolutions within a compressed video. The keyframe is generally at the beginning or end of an action~\cite{keyframe_1}. In between two keyframes are the P-frames and B-frames. P-frame, namely the predicted picture, holds only the variations between the current image frame and the closest preceding I-frame or P-frame in a video, occupying much less disk space than a keyframe. B-frame, the bidirectional predicted picture, is encoded from an interpolation of the closest succeeding and preceding image frames, either I-frame or P-frame, requiring the least amount of disk space~\cite{keyframe_2, Krishna2020I}. Other compression formats such as MPEG-4 and JPEG follow the same rule as H.264 except for the absence of B-frame. In Deepfake detection, the image frame extraction process suffers information loss of the P-frames and B-frames during image frame reconstruction in video decoding. Considering that the hyper-realistic Deepfake videos contain few fake cues that can be easily discovered, it is significant to collect as many keyframes from the video when detecting a fake face.

\subsection{Convolutional Local Feature Extraction}
\label{sec:met_conv_feat_extr}

We extract the local features from the input keyframe face images using a stack of CNNs. Specifically, the convolutional local feature extraction module contains 17 convolutional layers, each with a kernel size of three, and the stride and padding values are set to one. In particular, each convolutional layer is followed by batch normalization and a GELU activation function for feature normalization and non-linearity, respectively. The stack of CNNs is distributed into five, each followed by a max-pooling layer with a kernel size of two and stride value of two to concentrate on the dominant features in the input image. Except for the first group of CNNs that raises the number of input image channels from 3 to 32, the first convolutional layer of each group doubles the number of input channels. In addition, each max-pooling layer reduces the intermediate image feature dimensions by half. As a result, the convolutional local feature extraction module extracts a local feature with dimension $512 \times 7 \times 7$ by inputting an image with dimension $3 \times 224 \times 224$. The extracted local features are then fed into the Transformer part for global feature learning and relation analysis. 

\subsection{Deep Transformer with Convolutional Pooling and Re-attention}

The Transformer architecture~\cite{Vaswani2017Attention} is firstly introduced with the innovative multi-head self-attention mechanism, which learns global features and relations with low costs. In this section, inspired by the scheme of pooling vision Transformer~\cite{Byeongho2021Rethinking}, we boost its learning ability with depth-wise separable convolutional projection and deep re-attention to analyze the global features and relations within the input image.

\subsubsection{Convolutional Pooling Transformer}

The success of the Transformer architecture in the Natural Language Processing (NLP) domain is attributed to the autoencoder design and the multi-head self-attention mechanism. While the auto-encoder structure plays the role to fit the most classic NLP tasks, multi-head self-attention is the one that learns global features at a low cost. In specific, a Transformer includes various Transformer blocks, each containing a multi-head self-attention section followed by residual normalization and feedforward layers. In each Transformer block, three learnable linear projections $\mathbf{W}_Q$, $\mathbf{W}_K$, and $\mathbf{W}_V$ are performed on the input $\mathbf{X}$ to generate Query ($\mathbf{Q}$), Key ($\mathbf{K}$), and Value ($\mathbf{V}$) matrices by
\begin{align}
\mathbf{Q}, \mathbf{K}, \mathbf{V} = \mathbf{XW}_Q, \mathbf{XW}_K, \mathbf{XW}_V.
\label{eq1}
\end{align}
Then, the $\mathbf{Q}$, $\mathbf{K}$, and $\mathbf{V}$ matrices are arranged for self-attention computation by
\begin{align}
\textrm{Attention}(\mathbf{Q}, \mathbf{K}, \mathbf{V}) = \textrm{softmax}(\frac{\mathbf{QK}^{\textrm{T}}}{\sqrt{d_K}})\mathbf{V}, 
\label{eq2}
\end{align}
where $d_K$ is the dimension of each entry vector in $\mathbf{K}$.

\begin{figure}
\centering
\includegraphics[width=\columnwidth]{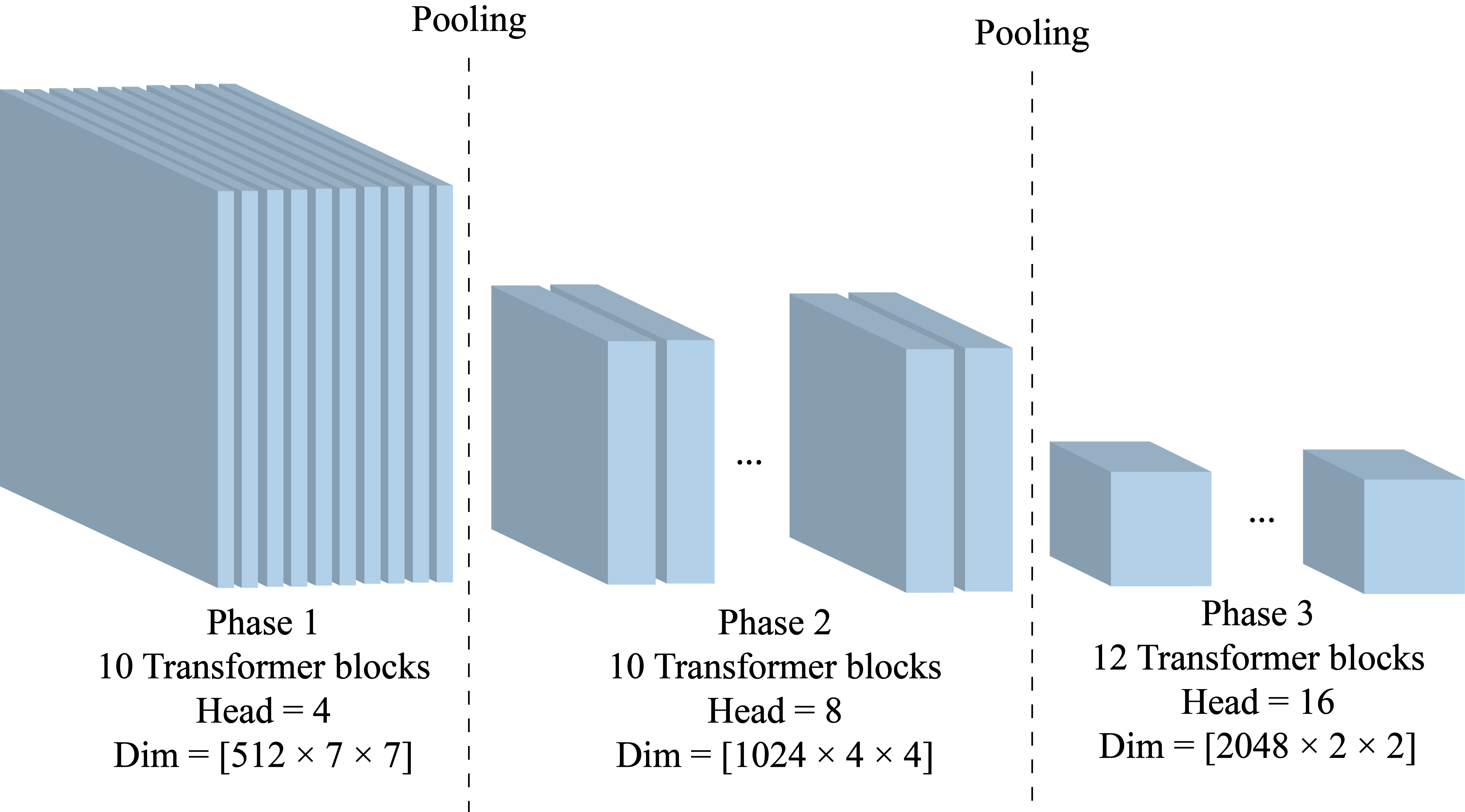}
\caption{Framework of the Transformer with convolutional pooling. The three phases are partitioned by two convolutional pooling layers, resulting in doubling the number of channels and reducing half of the feature map dimensions.}
\label{fig4}
\end{figure}

In the existing efforts, a convolutional implementation of the attention mechanism with a kernel size of one fails to study the relation between any two feature patches within an entire feature map. Meanwhile, the functionality is similar to a fully connected layer except for the modification of the number of channels. On the other hand, a CNN with a kernel size that is the same as the feature map dimension considers all feature patches and their relations by the convolutional computation. However, such an operation downscales the feature map to the dimension of one, which significantly discards determinant features and deteriorates the model performance. In our proposed approach, we apply a pooling Transformer to fit feature dimensions for image analysis. The self-attention mechanism makes the distance between any two locations in the feature map one and achieves global feature relation learning at a low cost. A multi-head self-attention is an ensemble of self-attentions, each with a unique group of learnable linear projections for $\mathbf{Q}$, $\mathbf{K}$, and $\mathbf{V}$ computation, offering multiple perspectives of global feature learning. In particular, for a squared input image fed to the convolutional local feature extraction module, a feature map maintaining the original relative position information with dimension $w \times w \times c$ is derived, where $w$ is the width of the extracted feature map and $c$ is the number of channels. The feature map can be regarded as $w^2$ feature patches in $c$ different channels of perspectives. The $\mathbf{Q}$, $\mathbf{K}$, and $\mathbf{V}$ matrices are then generated and flattened to the shape of $w^2 \times c$. We append the special learnable CLS token~\cite{Vaswani2017Attention} for classification purposes and split the channels regarding $h$ attention heads, which devotes the result with dimension $(w^2 + 1) \times h e$, where $c = h e$ and each of the $w^2 + 1$ rows represents the embedded feature information of a feature patch. In the self-attention calculation for each attention head, the multiplication of $\mathbf{Q}$ and $\mathbf{K}$ projected from the input $\mathbf{X}$ devotes the following matrix,
\begin{align}
\mathbf{QK}=
\begin{bmatrix}
\mathbf{q}_1\mathbf{K}, & \mathbf{q}_2\mathbf{K}, & ..., & \mathbf{q}_{w^2 + 1}\mathbf{K}
\end{bmatrix},
\label{eq3}
\end{align}
where $\mathbf{K} = \begin{bmatrix}\mathbf{k}_1, & \mathbf{k}_2, & ..., & \mathbf{k}_{w^2 + 1}\end{bmatrix}$. The entries on the $\mathbf{QK}$ matrix diagonal represent the self-relations for each feature patch, and the rest entries represent the relative relations between any two different feature patches. Omitting the constant computations and the softmax function, a multiplication with $\mathbf{V}$ further enhances the relation learning within the input image and derives the following matrix with dimension $(w^2 + 1) \times e$,
\begin{align}
\mathbf{QKV}=
\begin{bmatrix}
\mathbf{q}_1\mathbf{KV}, & \mathbf{q}_2\mathbf{KV}, & ..., & \mathbf{q}_{(w^2+1)}\mathbf{KV}
\end{bmatrix},
\label{eq4}
\end{align}
where $\mathbf{V} = \begin{bmatrix}\mathbf{v}_1, & \mathbf{v}_2, & ..., & \mathbf{v}_e\end{bmatrix}$ and each entry of $\mathbf{V}$ is the entire feature map at one of the $e$ feature channels. As a result, relations between any two feature patch positions within an input image are studied and applied to assist Deepfake detection.

Moreover, since the self-attention computation breaks the relative positional layout of the feature patches, a positional embedding is introduced to maintain meaningful relations between feature patches, and we apply sinusoidal position encoding to each feature patch after convolutional local feature extraction by
\begin{align}
\textrm{PE}(p, 2i) = \sin{(\frac{p}{\lambda^{\frac{2i}{d_{\textrm{model}}}}})},
\label{eq5}
\end{align}
and
\begin{align}
\textrm{PE}(p, 2i + 1) = \cos{(\frac{p}{\lambda^{\frac{2i}{d_{\textrm{model}}}}})},
\label{eq6}
\end{align}
where $\lambda$ is a constant controlling the sinusoidal cycle speed, $p$ is the position of the feature patch, $d_\textrm{model}$ is the feature patch dimension, and $i$ is the real-time running dimension in range $(0, d_\textrm{model})$.

\begin{figure}
\centering
\includegraphics[width=\columnwidth]{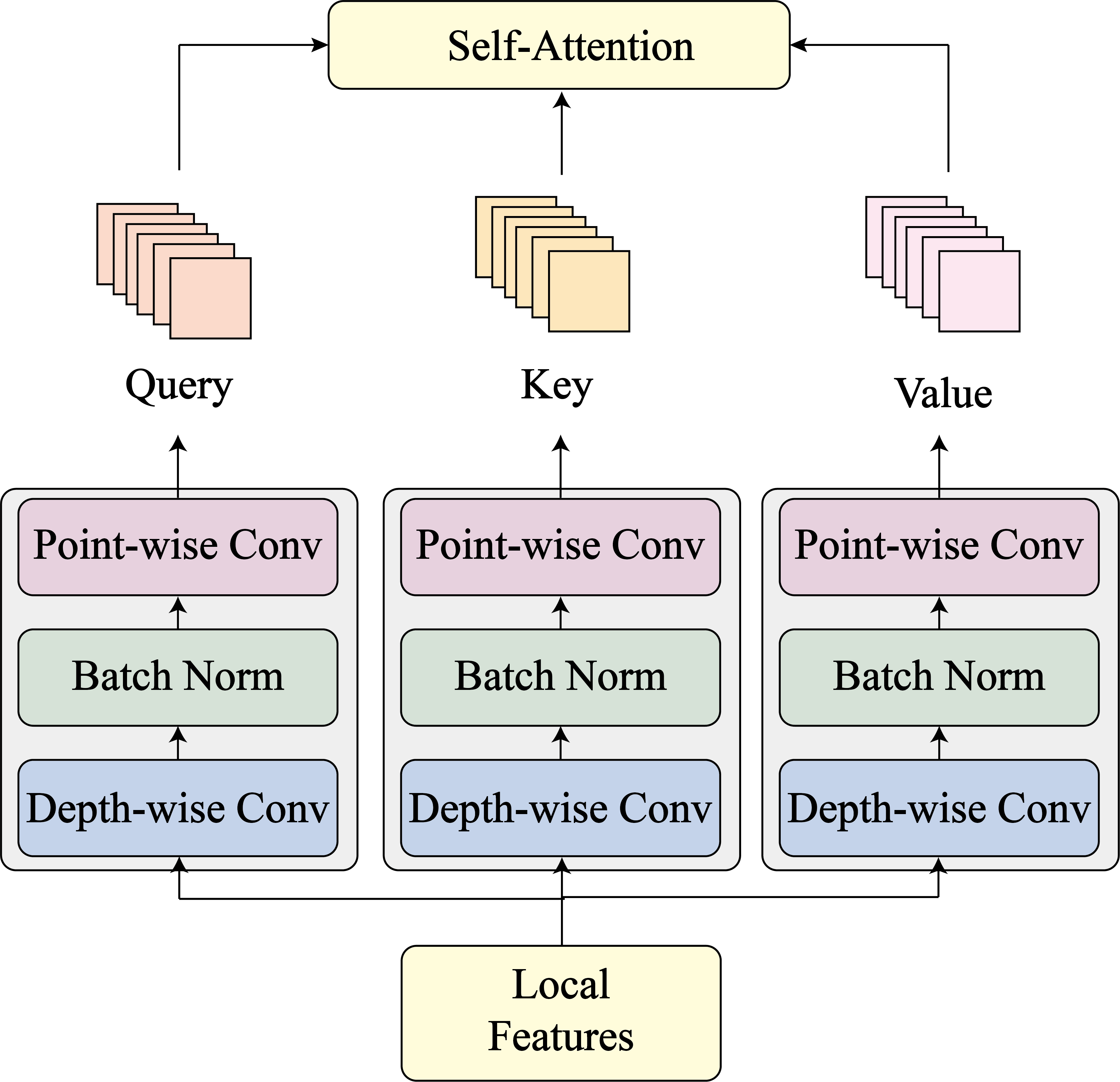}
\caption{Exhibition of the convolutional projection for $\mathbf{Q}$, $\mathbf{K}$, and $\mathbf{V}$ in each Transformer block. }
\label{fig5}
\end{figure}

The role of pooling is to adjust the receptive field, and the interaction within feature patches increases as the number of feature patches decreases after the pooling operation. In detail, the Transformer blocks are phased into three, with a convolutional pooling between every two consecutive phases. The convolutional pooling (as shown in Fig.~\ref{fig4}) is implemented as a convolutional layer with non-linearity and takes charge of refining dominant features. In particular, each convolutional pooling layer doubles the number of feature channels and reduces the feature dimension by half.

Considering the advantages of CNNs over linear projections on image analysis, we further introduce the depth-wise separable convolutions into the generation process of $\mathbf{Q}$, $\mathbf{K}$, and $\mathbf{V}$ matrices to enrich image feature learning for multi-head self-attention computation. In detail, we replace the three fully connected layers for the generation of $\mathbf{Q}$, $\mathbf{K}$, and $\mathbf{V}$, each with a depth-wise separable convolution followed by batch normalization when projecting the input features in each Transformer block. To maintain stable model architecture, the separable convolution is set to only perform projections while maintaining the feature dimensions as being extracted from the local feature extraction module. The workflow of the separable convolutional projection is shown in Fig.~\ref{fig5}. The depth-wise separable convolution is to avoid the possible overfitting and time-consuming problem due to the large number of parameters in a standard CNN~\cite{Chollet2017Xception}. Specifically, the depth-wise separable convolution divides a standard CNN into a depth-wise convolution and a point-wise convolution. The former is a spatial convolution that only modifies the feature map dimension with the channels unchanged, while the latter only changes the number of channels. By adopting the depth-wise separable convolution for matrix projections, the proposed model can analyze richer image features. 

\begin{table*}
\begin{center}
\caption{Summary of the attending datasets in this study. Information includes the publication date, manipulation techniques, number of real and fake videos, and video compression format. }
\resizebox{\textwidth}{!}{
\begin{tabular}{lllll}
\toprule\noalign{\smallskip}
Dataset  & Publish Time & Manipulation Techniques                                                                                                   & \# Real / Fake & Format \\ 
\noalign{\smallskip}\midrule\noalign{\smallskip}
FF++~\cite{Rossler2019FaceForensics}     & Jan. 2019         & DF, F2F, FS, NT                                                                                                 & 1,000 / 4,000    & H.264  \\
Celeb-DF~\cite{CelebDF2020} & Nov. 2019         & Improved Deepfake                                                                                               & 590 / 5,639     & H.264  \\
DF-1.0~\cite{jiang2020deeperforensics1}   & Jan. 2020         & DF-VAE with Seven Perturbations & 50,000 / 10,000  & JPEG   \\
DFDC~\cite{dolhansky2020deepfake}     & June 2020 & DF-128, DF-256, MM/NN, NTH, FSGAN, StyleGAN, Refinement, Audio Swaps & 23,654 / 104,500 & H.264  \\
\noalign{\smallskip}\bottomrule
\end{tabular}
}
\label{tab1}
\end{center}
\end{table*}

\subsubsection{Deep Re-attention}
A deep model with a mass of Transformer blocks is necessary for complex image tasks. However, a deep ViT model empirically suffers from the attention collapse issue, such that the attention maps tend to be overly similar, and obvious performance decline is observed as the model goes deeper~\cite{Daquan2021DeepViT}. In other words, the multi-head self-attention mechanism begins to taper off in feature learning efficacy as the model further proceeds. Therefore, in this study, we construct a deep pooling Transformer with a total of 24 Transformer blocks in three phases and employ the re-attention technique~\cite{Daquan2021DeepViT} with a learnable transformation matrix $\mathbf{\Theta}$ to retain attention map diversity. Specifically, observing that the attention maps from different attention heads are not affected by the attention collapse, the re-attention technique resorts to cross-head communication by
\begin{align}
\textrm{Re-Attention}(\mathbf{Q}, \mathbf{K}, \mathbf{V}) = \mathbf{\Theta}^{\textrm{T}}\textrm{softmax}(\frac{\mathbf{QK}^{\textrm{T}}}{\sqrt{d_K}})\mathbf{V},
\label{eq7}
\end{align}
the multiplication of the learnable parameter $\mathbf{\Theta}$ and the self-attention map along the head dimension. The attention map diversity is improved with the unique learnable transformation matrix $\mathbf{\Theta}$ for the multi-head self-attention within each Transformer block. Consequently, richer features and relations can be learned and boost the Deepfake detection performance.

\subsubsection{Binary Classification}
Following the convention for binary classification, we take the learnable CLS token from the output after the last Transformer block and pass it through a fully connected layer followed by the softmax layer for Deepfake detection prediction on the input face image frame. The model is tuned with the cross entropy loss function by
\begin{align}
L_{\textrm{CE}}=-\Sigma^{2}_{i=1}t_i\log{p_i},
\label{eq8}
\end{align}
where $t_i$ is the groundtruth value and $p_i$ is the softmax prediction for class $i$.

\section{Experiments}

In this section, we first detail the implementation of the model training process. Then, we describe the selected Deepfake datasets for training and testing. Thereafter, we conduct experiments to evaluate the model performance on the FaceForensics++ (FF++) dataset and several other Deepfake datasets. Finally, we propose several ablation studies to evaluate the importance of keyframes, the performance of each model component, and the effectiveness of different model depths. 

\subsection{Implementation Details}

We used the DLIB library~\cite{dlib} to detect faces for all image frames and saved the aligned facial images with a size of $224 \times 224$. The depth of the pooling Transformer is set to 24 of three phases with 8, 8, and 8 Transformer blocks and 4, 8, and 16 attention heads, respectively. The sinusoidal cycle constant $\lambda$ is set to 10,000. Adam optimizer with a learning rate of $1e-4$ and weight decay of $1e-4$ is used for optimization. The models are trained on the Tesla V100 GPU with batch size 32. All models are trained and tested at frame level upon the extracted facial images. 

\subsection{Datasets}
FaceForensics++ (FF++)~\cite{Rossler2019FaceForensics} is currently the most widely adopted dataset in the existing Deepfake detection studies. It includes 1,000 original pristine videos collected from YouTube and four subsets each containing 1,000 fake videos synthesized from the 1,000 original real ones using a unique facial manipulation technique. The four facial manipulations are known as FaceSwap (FS), Deepfakes (DF), Face2Face (F2F)~\cite{Thies2016Face2Face}, and NeuralTextures (NT)~\cite{Thies2019Deferred}. In general, the FF++ dataset provides an official split with a ratio of 720:140:140 for training, validation, and testing videos. Three qualities have been released, namely, Raw, HQ (c23), and LQ (c40), where the latter two are compressed using the H.264 codec with different compression levels. We adopted the HQ (c23) version dataset, which is similar to the real-world Deepfake, and followed the official split ratio to execute image frame extraction and face extraction using the DLIB library. 

As discussed in section~\ref{sec:met_keyframe} and following the early approach for keyframe extraction~\cite{spider2013Naghmeh}, we first derived all keyframe faces from the videos using FFmpeg~\cite{ffmpeg} and augmented\footnote{Random augmentations include random rotation, transposition, horizontal flipping, vertical flipping, etc.} the real keyframe faces for a balanced keyframe dataset. Due to the limited amount, it is insufficient to train the model to converge with keyframes only. Therefore, we further randomly extracted sufficient normal frames from each video. Specifically, to maintain a balanced dataset between the real and fake faces, we extracted three times more normal frame faces from each real video than the fake one. In this study, the proposed model is trained on the FF++ training set and evaluated through the within-dataset test on the FF++ testing set with a mixture of keyframe and normal frame faces. Meanwhile, we collected various popular Deepfake benchmark datasets for extensive experiments on cross-dataset evaluation, namely, Deepfake Detection Challenge (DFDC)~\cite{dolhansky2020deepfake}, Celeb-DF~\cite{CelebDF2020}, and DeeperForensics-1.0 (DF-1.0)~\cite{jiang2020deeperforensics1}. A summary of the attending datasets is listed in Table~\ref{tab1}.

\begin{table*}
\begin{center}
\caption{Quantitative frame-level ACC and AUC performance comparison on each testing set after training on the FF++ dataset. The best performance is marked as bold. }
\resizebox{\textwidth}{!}{
\begin{tabular}{lllllllll}
\toprule\noalign{\smallskip}
\multirow{3}{*}{Method} & \multicolumn{8}{c}{Test Dataset} \\
\noalign{\smallskip}\cmidrule{2-9}\noalign{\smallskip}
& \multicolumn{2}{c}{FF++} & \multicolumn{2}{c}{DFDC} & \multicolumn{2}{c}{Celeb-DF} & \multicolumn{2}{c}{DF-1.0} \\
\noalign{\smallskip}\cmidrule{2-9}\noalign{\smallskip}
& ACC & AUC & ACC & AUC & ACC & AUC & ACC & AUC \\
\noalign{\smallskip}\midrule\noalign{\smallskip}
MesoNet~\cite{Afchar2018mesonet} & 61.03\% & 58.13\% & 50.02\% & 50.16\% & 36.73\% & 50.01\% & 50.05\% & 50.21\% \\
Capsule~\cite{nguyen2019use} & 76.40\% & 83.44\% & 51.30\% & 56.16\% & 61.96\% & 59.93\% & 59.29\% & 61.46\% \\
FFD~\cite{Dang2020on} & 82.29\% & 82.48\% & 59.44\% & 59.47\% & 46.19\% & 55.86\% & 53.69\% & 53.81\% \\
CViT~\cite{Deressa2021Deepfake} & 83.05\% & 91.08\% & 60.76\% & 67.43\% & 53.26\% & 63.60\% & 54.97\% & 58.52\% \\
MAT~\cite{Zhao2021Multi-Attentional} & 87.50\% & 94.85\% & 63.16\% & 69.56\% & 44.78\% & 57.20\% & 56.90\% & 61.72\% \\
SRM~\cite{Luo2021Generalizing} & 88.17\% & 94.93\% & 59.93\% & 64.80\% & 52.95\% & 60.90\% & 55.83\% & 62.54\% \\
Xception~\cite{Chollet2017Xception} & 90.08\% & 96.51\% & 58.77\% & 66.95\% & 54.24\% & 65.86\% & 54.76\% & 67.03\% \\
Our Approach & \textbf{92.11\%} & \textbf{97.66\%} & \textbf{65.76\%} & \textbf{73.68\%} & \textbf{63.27\%} & \textbf{72.43\%} & \textbf{62.46\%} & \textbf{78.19\%} \\
\noalign{\smallskip}\bottomrule
\end{tabular}
}
\label{tab2}
\end{center}
\end{table*}

Deepfake Detection Challenge (DFDC)~\cite{dolhansky2020deepfake} is one of the largest public Deepfake datasets~\cite{WildDeepfake2020, KoDF2021Kwon, ForgeryNet2021He} by far with 128,154 videos, which includes synthesized videos using eight different facial manipulation methods as listed in Table~\ref{tab1}. However, its large quantity has caused storage issues in data preprocessing such that the existing approaches have been seldomly attempted on the whole dataset. In this study, we acquired the entire DFDC dataset and randomly shuffled 10\% of the whole dataset for cross-dataset evaluation. 

Celeb-DF~\cite{CelebDF2020} is one of the most challenging Deepfake detection datasets because of the high quality and high difficulty of its official testing set with 518 videos. It has failed most of the existing baselines at a time, even in the within-dataset performance on the testing set. We considered the Celeb-DF official testing dataset with 518 videos in the extensive experiments and evaluated the robustness of our approach against the existing baseline models. It is worth noting that the official testing set is imbalanced with a ratio of 178:340 for real and fake.

DeeperForensics-1.0 (DF-1.0)~\cite{jiang2020deeperforensics1} is the first large dataset that has been manually added deliberate distortions and perturbations to the clean Deepfake videos. The additional noise aims to imitate real-world scenarios and has brought new challenges to the Deepfake detection task. The dataset contains a total of 60,000 videos with a ratio of 5:1 for real and fake. The fake videos are generated based on the real source videos, and seven types and five levels of distortions and perturbations are added accordingly. We followed the official split scheme and performed Deepfake detection on the 1,000 testing set videos with mixed levels of random distortions and perturbations added. To keep a balanced testing dataset, we randomly sampled the same amount of real videos in the experiment. 

Since the above three datasets are rather recently published with high-quality videos, we successfully extracted sufficient numbers of keyframes to establish each testing set for cross-evaluation. Unless an imbalanced official testing set is given (Celeb-DF), we constructed balanced testing sets for all experiments. Besides, all comparative experiments are conducted on the same datasets utilized by our proposed model for complete fairness.

\begin{figure*}
\centering
\includegraphics[width=\textwidth]{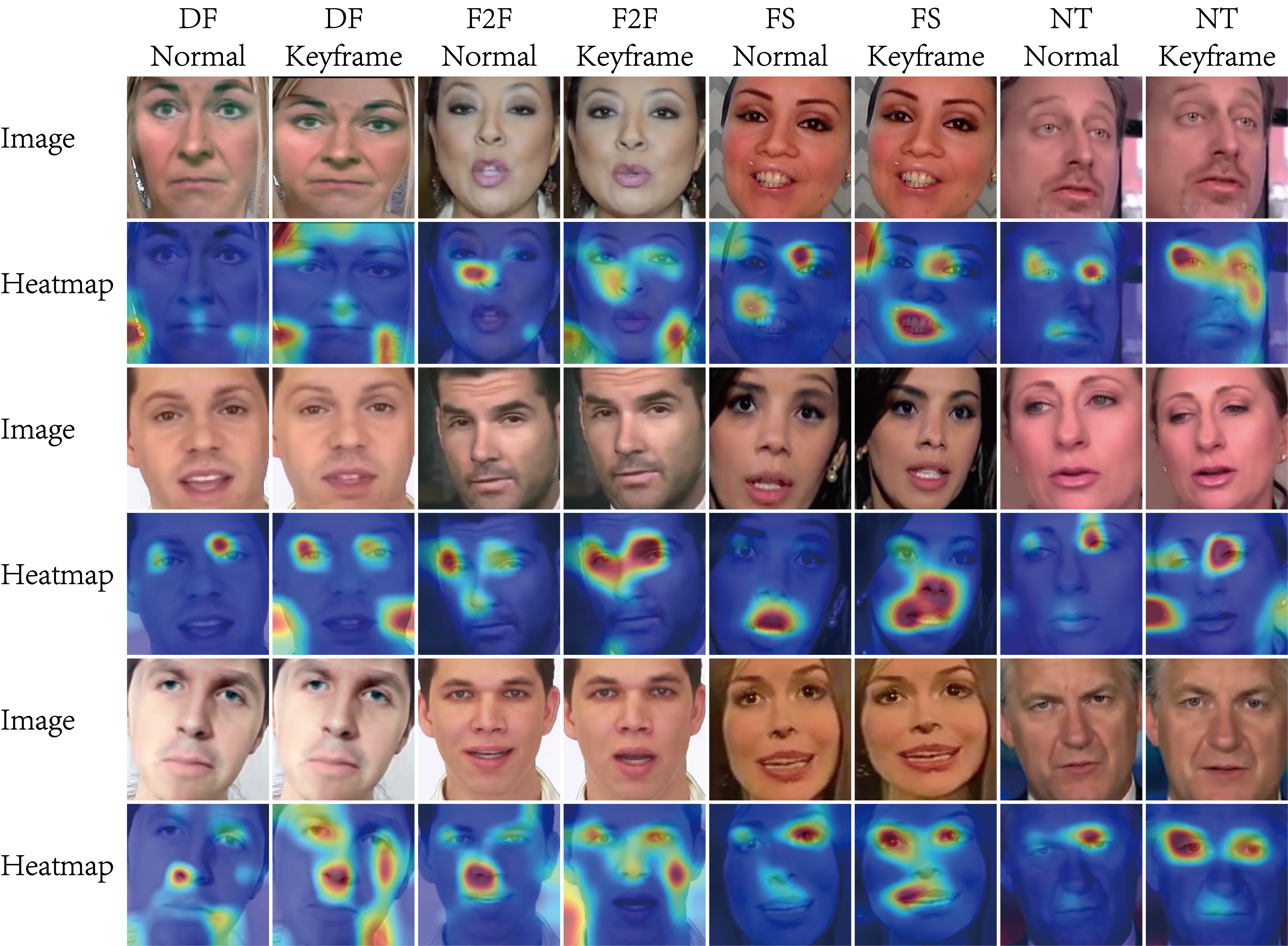}
\caption{Feature heatmap of pairs of keyframes and normal frames. Every two columns display three pairs of feature heatmaps of each manipulation technique in FF++. Each pair includes a keyframe and a normal frame from the same video, along with their feature heatmaps. A noticeable feature gap can be observed within each pair. }
\label{fig6}
\end{figure*}

\begin{table}
\begin{center}
\caption{Frame-level ACC and AUC Performance comparison between with and without keyframes in training and testing. The models are trained on the FF++ dataset. }
\resizebox{\columnwidth}{!}{
\begin{tabular}{llll}
\toprule\noalign{\smallskip}
\multirow{2}{*}{Test Data} & & \multicolumn{2}{c}{Trained on FF++} \\
\noalign{\smallskip}\cmidrule{3-4}\noalign{\smallskip}
 &  & w/o keyframes & w/ keyframes \\
\noalign{\smallskip}\midrule\noalign{\smallskip}
\multirow{2}{*}{FF++} & ACC & 83.74\% & 92.11\% \\
 & AUC & 91.44\% & 97.66\% \\
\multirow{2}{*}{DFDC} & ACC & 61.41\% & 65.76\% \\
 & AUC & 68.72\% & 73.68\% \\
\multirow{2}{*}{Celeb-DF} & ACC & 52.72\% & 63.27\% \\
 & AUC & 64.28\% & 72.43\% \\
\multirow{2}{*}{DF-1.0} & ACC & 55.94\% & 62.46\% \\
 & AUC & 68.00\% & 78.19\% \\
\noalign{\smallskip}\bottomrule
\end{tabular}%
}
\label{tab3}
\end{center}
\end{table}

\begin{table}
\begin{center}
\caption{Statistics of the training sets with four different settings. K: keyframe faces solely; $\textrm{K}_{\textrm{aug}}$: keyframe faces with data augmentation; N: normal frame faces solely; $\textrm{K} + \textrm{N}$: the combination of key and normal frame faces. }
\resizebox{\columnwidth}{!}{
\begin{tabular}{llll}
\toprule\noalign{\smallskip}
training data & real & fake & total \\
\noalign{\smallskip}\midrule\noalign{\smallskip}
K & 7,068 & 6,548 & 13,616 \\
$\textrm{K}_{\textrm{aug}}$ & 21,204 & 19,644 & 40,848 \\
N & 19,540 & 19,531 & 39,071 \\
$\textrm{K} + \textrm{N}$ & 26,608 & 26,085 & 52,693 \\
\noalign{\smallskip}\bottomrule
\end{tabular}
}
\label{tab4}
\end{center}
\end{table}

\subsection{Performance Evaluation}
\label{sec:exp_perf_eval}

We first evaluated our approach on the FF++ testing set after training and then examined the model transferability on DFDC, Celeb-DF, and DF-1.0 testing sets. We also performed comparative tests against the state-of-the-art Deepfake detection approaches to justify the performance of our model. Both the proposed model and the comparative models are trained on the FF++ dataset.

For fairness, we only adopted the state-of-the-art baselines with publicly available source code and trained and tested each comparative method following the corresponding default optimal parameter settings on the same datasets we used for our model evaluation. After training, our proposed model achieves a 92.11\% accuracy (ACC) and a 97.66\% area under the receiver operating characteristic (ROC) curve (AUC) score on the FF++ testing set. As Table~\ref{tab2} columns two and three shown, although some comparative approaches have derived competitive performance, our proposed model has outperformed all baseline models in the comparative experiment on FF++.

We further conducted extensive experiments on cross-dataset tests for DFDC, Celeb-DF, and DF-1.0. As Table~\ref{tab2} columns four to nine shown, our proposed Deepfake detection approach outperforms all state-of-the-art baselines in the comparative tests for the cross-dataset experiment. It can be concluded that the CNN-based approaches such as MesoNet~\cite{Afchar2018mesonet}, Capsule~\cite{nguyen2019use}, and FFD~\cite{Dang2020on} mainly focus on the local features within the face images, which lack global information for further enhancement and thus have poor transferability when cross-evaluated on DFDC, Celeb-DF, and DF-1.0. It can also be observed that the Transformer based CViT~\cite{Deressa2021Deepfake} approach directly adopts the vision Transformer (ViT)~\cite{Alexey2021An} where the utilized vanilla Transformer is unstable on image analysis tasks. Besides, the recent state-of-the-art models MAT~\cite{Zhao2021Multi-Attentional} and SRM~\cite{Luo2021Generalizing} have shown relatively competitive abilities for both within-dataset and cross-dataset evaluations compared to the early approaches. Nevertheless, the Xception~\cite{Chollet2017Xception} method which utilized the XceptionNet CNN backbone has derived surprisingly promising performance with even higher accuracies and AUC scores than the most recent state-of-the-art models. This might be due to the robustness of the perfect XceptionNet architecture design.

As a result, although the manipulation techniques in the testing sets are unseen during the training process for the cross-dataset evaluation, our proposed method is the only one that achieves over 60\% accuracy and 70\% AUC scores for all of DFDC, Celeb-DF, and DF-1.0 comparing to other approaches. According to the AUC metric, we observed that the high-quality Celeb-DF dataset was the most challenging one for most approaches. It is worth noting that our model also performs stably on the DF-1.0 dataset with random levels of distortions and perturbations, which means the model is robust against additional artificial noise. 

\begin{table*}
\begin{center}
\caption{Performance evaluation using different training and testing settings with the FF++ dataset. K: keyframe faces solely; $\textrm{K}_{\textrm{aug}}$: keyframe faces with data augmentation; N: normal frame faces solely; $\textrm{K} + \textrm{N}$: the combination of key and normal frame faces. }
\resizebox{\textwidth}{!}{
\begin{tabular}{lllllllll}
\toprule\noalign{\smallskip}
\multirow{3}{*}{Tested on FF++} & \multicolumn{8}{c}{Trained on FF++} \\
\noalign{\smallskip}\cmidrule{2-9}\noalign{\smallskip}
& \multicolumn{2}{c}{K} & \multicolumn{2}{c}{$\textrm{K}_{\textrm{aug}}$} & \multicolumn{2}{c}{N} & \multicolumn{2}{c}{$\textrm{K} + \textrm{N}$} \\
\noalign{\smallskip}\cmidrule{2-9}\noalign{\smallskip}
& ACC & AUC & ACC & AUC & ACC & AUC & ACC & AUC \\
\noalign{\smallskip}\midrule\noalign{\smallskip}
K & 82.03\% & 90.20\% & 89.41\% & 96.11\% & 82.48\% & 90.34\% & 92.09\% & 97.64\% \\
N & 72.66\% & 82.01\% & 83.54\% & 93.87\% & 84.18\% & 91.83\% & 92.12\% & 97.69\% \\
$\textrm{K} + \textrm{N}$ & 75.09\% & 84.23\% & 85.06\% & 94.35\% & 83.74\% & 91.44\% & 92.11\% & 97.66\% \\
\noalign{\smallskip}\bottomrule
\end{tabular}
}
\label{tab5}
\end{center}
\end{table*}

\subsection{Keyframe Analyses}

In this study, we emphasized the significant role that the keyframes play in video compression, and we claimed that our model studies richer features and relations from the keyframes than from the normal frames. In particular, we conducted ablation studies setting the variation in the attendance of keyframe faces during the training and testing process. Firstly, having the model architecture fixed, we enforced two training sessions with one adopting normal image frames and keyframes, and the other merely normal image frames. Models trained in the two sessions are evaluated on the same group of testing sets as introduced in section~\ref{sec:exp_perf_eval}. As shown in Table~\ref{tab3}, the model session trained with keyframes performs remarkably better than the one trained with only normal image frames. Specifically, the attendance of keyframes has boosted the model performance by 6.22\%, 4.96\%, 8.15\%, and 10.19\% when tested on FF++, DFDC, Celeb-DF, and DF-1.0, respectively. The results have proved the superiority of keyframes over normal frames in feature richness.

Moreover, we employed the gradient-based visualization approach, the Grad-CAM~\cite{Selvaraju2017Grad-CAM}, for feature heatmap visualization upon our proposed model. We visualized the facial features extracted by the proposed model from randomly selected pairs of keyframes and normal frames. We then displayed four pairs of keyframes and normal frames for each facial manipulation technique in the FF++ dataset along with the heatmaps of their extracted features (Fig.~\ref{fig6}), where each pair includes a keyframe and a normal frame face derived from the same video. Specifically, every two rows exhibit pairs of facial images and feature heatmaps, and every two columns illustrate the pairs of keyframes and normal frames for each manipulation technique. For a feature heatmap, the hotter (colored with deep red) an area is, the more features are extracted by the model. It can be easily observed that although faces from the same video contain `hot' features at similar locations, the keyframe faces are generally `hotter' with larger red area than the normal frames. This means that the keyframe faces have relatively more decisive features extracted by the proposed model, which helps derive better detection performance. Consequently, the displayed features from any pair of faces are sufficient to conclude the advantage of keyframes over normal frames in Deepfake detection.

To imitate the possible real-life cases, we further probed the importance of keyframes by training and testing with different combinations of keyframes and normal frames. In detail, we trained the proposed model with four different settings, namely, keyframes (K), keyframes with data augmentation ($\textrm{K}_{\textrm{aug}}$), normal frames (N), and the combination of key and normal frames ($\textrm{K} + \textrm{N}$). The reason for having the keyframes with data augmentation setting is the lack of sufficient FF++ keyframe data for training the model. Statistics of the four training data settings are listed in Table~\ref{tab4}, and we kept the datasets always roughly balanced for real and fake. Models trained with the above four settings are each evaluated on FF++ with the settings keyframes (K), normal frames (N), and the combination of key and normal frames ($\textrm{K} + \textrm{N}$). The results are exhibited in Table~\ref{tab5}.

\begin{table*}
\begin{center}
\caption{Ablation study on module cumulation performance. The components pooling, convolutional (conv.) projection, and re-attention are cumulatively added starting from a vanilla Transformer. }
\resizebox{\textwidth}{!}{
\begin{tabular}{lllllllll}
\toprule\noalign{\smallskip}
\multirow{3}{*}{Method} & \multicolumn{8}{c}{Test Dataset} \\
\noalign{\smallskip}\cmidrule{2-9}\noalign{\smallskip}
& \multicolumn{2}{c}{FF++} & \multicolumn{2}{c}{DFDC} & \multicolumn{2}{c}{Celeb-DF} & \multicolumn{2}{c}{DF-1.0} \\
\noalign{\smallskip}\cmidrule{2-9}\noalign{\smallskip}
& ACC & AUC & ACC & AUC & ACC & AUC & ACC & AUC \\
\noalign{\smallskip}\midrule\noalign{\smallskip}
Vanilla Transformer & 83.21\% & 91.60\% & 62.62\% & 66.90\% & 52.71\% & 62.18\% & 52.09\% & 60.97\% \\
+ Pooling & 87.38\% & 94.63\% & 62.73\% & 69.36\% & 54.28\% & 63.90\% & 54.44\% & 64.47\% \\
+ Conv. Projection & 91.45\% & 97.36\% & 62.92\% & 70.34\% & 60.29\% & 68.26\% & 57.86\% & 69.17\% \\
+ Re-attention & 92.11\% & 97.66\% & 65.76\% & 73.68\% & 63.27\% & 72.43\% & 62.46\% & 78.19\% \\
\noalign{\smallskip}\bottomrule
\end{tabular}
}
\label{tab6}
\end{center}
\end{table*}

\begin{table*}
\begin{center}
\caption{Ablation study on model depth. }
\resizebox{\textwidth}{!}{
\begin{tabular}{lllllllll}
\toprule\noalign{\smallskip}
\multirow{3}{*}{Model Depth} & \multicolumn{8}{c}{Test Dataset} \\
\noalign{\smallskip}\cmidrule{2-9}\noalign{\smallskip}
& \multicolumn{2}{c}{FF++} & \multicolumn{2}{c}{DFDC} & \multicolumn{2}{c}{Celeb-DF} & \multicolumn{2}{c}{DF-1.0} \\
\noalign{\smallskip}\cmidrule{2-9}\noalign{\smallskip}
& ACC & AUC & ACC & AUC & ACC & AUC & ACC & AUC \\
\noalign{\smallskip}\midrule\noalign{\smallskip}
$[\textrm{4, 4, 6}]$ & 89.70\% & 96.52\% & 61.12\% & 67.88\% & 53.09\% & 64.92\% & 56.90\% & 66.91\% \\
$[\textrm{8, 8, 8}]$ & 92.11\% & 97.66\% & 65.76\% & 73.68\% & 63.27\% & 72.43\% & 62.46\% & 78.19\% \\
$[\textrm{10, 10, 12}]$ & 92.78\% & 98.03\% & 64.85\% & 71.30\% & 63.37\% & 70.88\% & 58.61\% & 73.01\% \\
\noalign{\smallskip}\bottomrule
\end{tabular}
}
\label{tab7}
\end{center}
\end{table*}

It can be observed that training with sole keyframes derives unsatisfactory performance on all testing sets due to the issue of insufficient amount of training data. On the contrary, the model trained with an augmented keyframe dataset achieves significantly better performance, and the highest accuracy and AUC score are derived on the keyframe testing set as expected. As for the model trained with solely normal frames, although evaluating on the normal frame testing set outperforms that on the keyframe testing set, the overall performance is relatively weak due to the poor feature richness issue. Expectedly, training with the combination of key and normal frames has devoted the best performance on all three testing sets. 

In real life, it is unpredictable and uncontrollable whether a candidate Deepfake image is a key or normal frame. With the performance reported in Table~\ref{tab5}, it can be concluded that our proposed model after training with the combination of key and normal frames can achieve satisfactory detection performance on both key and normal frame Deepfake candidate faces. To be more specific, training with the combination of key and normal frames maintains robust model detection performance, in other words, the optimal performance in this study, regardless of the types of testing image frames. 

\subsection{Model Component Analysis}

To evaluate the contribution of each module, we conducted an ablation study on different components of the proposed model. In specific, having the local feature extraction module fixed, every other module is cumulatively added starting from a vanilla Transformer. As exhibited in Table~\ref{tab6}, pooling, convolutional projection, and re-attention modules are added step by step, and each module consistently boosts the detection performance as expected. 

It can be observed that the pooling module and the convolutional projection idea improve the performance evenly with around 3\% with respect to the AUC score when tested on FF++. While for the cross-dataset evaluation, the pooling module is more helpful on DFDC than on Celeb-DF, and the convolutional projection is more effective on Celeb-DF than on DFDC. Meanwhile, the two modules both boost the performance on DF-1.0 by around 4\% of AUC scores. Besides, re-attention is more instrumental in cross-dataset performance by adding diversity to the attention maps. In specific, it only gains less than 1\% AUC score improvement on FF++, which is relatively trivial compared to the performance of cross-dataset evaluation. It is also worth noting that each of the three components has brought significant performance enhancement when tested on DF-1.0. This informs that the components are still powerful and helpful with reliable robustness even when the candidate face image contains artificial noise.

\subsection{Model Depth Analysis}

In this study, the model depth of the pooling Transformer is set to 24 with three phases having the pooling layers in between every two consecutive phases. In this section, we conducted an ablation study on the effectiveness of various model depths. Specifically, we introduced three experiments with different model depths and evaluated the performance accordingly. The three experiments are conducted upon the model with depths [4, 4, 6], [8, 8, 8], and [10, 10, 12] for the three phases and having the total depths of 14, 24, and 32 from shallow to deep, respectively.

As Table~\ref{tab7} demonstrated, it can be concluded that deeper detection models do not always derive better performance than shallow ones. To take a closer look, the detection performance on FF++ consistently increases as the model depth becomes larger. However, the rate of increase is relatively trivial with around a 1\% AUC score gap between the adjacent experiments on FF++. On the contrary, a model with depth 24 achieves the best cross-dataset performance with 5.8\%, 7.51\%, and 11.28\% advantages on DFDC, Celeb-DF, and DF-1.0, respectively, over the shallow model with depth 14 concerning the AUC score. Meanwhile, it can be observed that the model starts to overfit on FF++ with performance damping by 2.38\%, 1.55\%, and 5.18\% AUC scores on DFDC, Celeb-DF, and DF-1.0, respectively, as the model depth goes deeper from 24 to 32. Thus, model depth 24 is the optimal setting in this study.

\section{Conclusion}
As the real-life Deepfake videos circulating on the Internet become more realistic and difficult to be identified, Deepfake detection models are facing considerable challenges, especially the transferability to the unseen Deepfake contents. In this paper, we present a novel deep convolutional Transformer using convolutional pooling and re-attention techniques for facial feature learning both locally and globally in Deepfake detection. We also demonstrate the significance of the barely discussed image keyframes in image feature learning as normal image frames suffer from information loss with common video compression. Compared with the existing Deepfake detection baselines, our approach achieves state-of-the-art performance in both within- and cross-dataset experiments with robust transferability. Furthermore, the visualization of the learned facial features has shown richer information obtained from the keyframes over the normal frames. Therefore, thoroughly exploring the image keyframes and studying the decisive features and relations both locally and globally can facilitate further performance enhancement in Deepfake detection.

Despite satisfactory experiment performance on Deepfake detection tasks, we are aware of the potential limitations. While the keyframes are of high image quality and contain more dominant features, they may be utilized in Deepfake generation. If the opponents intentionally raise the authenticity of Deepfake upon the keyframes, our proposed solution is highly likely to be fooled.


\bibliography{IEEEabrv, my_bib}

\bibliographystyle{IEEEtran}

\end{document}